\documentclass[11pt,a4paper]{article}
\usepackage[hyperref]{acl2019}
\usepackage{times}
\usepackage{latexsym}
\usepackage{amsmath}
\usepackage{amsfonts}
\usepackage{url}
\usepackage{bbm}
\usepackage{graphicx, wrapfig, graphics}
\usepackage{subfig}
\usepackage{tabulary}
\usepackage{multirow}
\usepackage{booktabs}
\usepackage{xcolor}
\usepackage{algorithm}
\usepackage[noend]{algpseudocode}
\DeclareMathAlphabet\EuRoman{U}{eur}{m}{n}
\SetMathAlphabet\EuRoman{bold}{U}{eur}{b}{n}
\algnewcommand\algorithmicinput{\textbf{INPUT:}}
\algnewcommand\algorithmicoutput{\textbf{OUTPUT:}}
\algnewcommand\And{\textbf{and}}
\algnewcommand\INPUT{\item[\algorithmicinput]}
\algnewcommand\OUTPUT{\item[\algorithmicoutput]}
\makeatletter
\def\BState{\State\hskip-\ALG@thistlm}
\definecolor{caribbeangreen}{rgb}{0.0, 0.8, 0.6}
\definecolor{brandeisblue}{rgb}{0.0, 0.44, 1.0}
\makeatother

\aclfinalcopy 
\newcommand{\suchthat}{\;\ifnum\currentgrouptype=16 \middle\fi|\;}
\newcommand{\rmx}{\mathrm{x}}
\newcommand{\rmy}{\mathrm{y}}

\newcommand{\green}{\color{caribbeangreen}}

\title{Decomposable Neural Paraphrase Generation}

\author{Zichao Li, Xin Jiang, Lifeng Shang, Qun Liu\\
Huawei Noah's Ark Lab\\
{\tt \{li.zichao, jiang.xin, shang.lifeng, qun.liu\}@huawei.com}
}

\date{}

\begin{document}
\maketitle

\begin{abstract}
Paraphrasing exists at different granularity levels, such as lexical level, phrasal level and sentential level. This paper presents Decomposable Neural Paraphrase Generator (DNPG), a Transformer-based model that can learn and generate paraphrases of a sentence at different levels of granularity in a disentangled way. Specifically, the model is composed of multiple encoders and decoders with different structures, each of which corresponds to a specific granularity. The empirical study shows that the decomposition mechanism of DNPG makes paraphrase generation more interpretable and controllable. Based on DNPG, we further develop an unsupervised domain adaptation method for paraphrase generation. Experimental results show that the proposed model achieves competitive in-domain performance compared to the state-of-the-art neural models, and significantly better performance when adapting to a new domain.

\end{abstract}

\section{Introduction}
Paraphrases are texts that convey the same meaning using different wording. Paraphrase generation is an important technique in natural language processing (NLP), which can be applied in various downstream tasks such as information retrieval, semantic parsing, and dialogue systems. Neural sequence-to-sequence (Seq2Seq) models have demonstrated the superior performances on generating paraphrases given a sentence~\citep{prakash2016neural,cao2017joint,li2017paraphrase,ma2018query}. All of the existing works learn to paraphrase by mapping a sequence to another, with each word processed and generated in a uniform way.

This work is motivated by a commonly observed phenomenon that the paraphrase of a sentence is usually composed of multiple paraphrasing patterns at different levels of granularity, e.g., from the lexical to phrasal to sentential levels. For instance, the following pair of paraphrases contains both the phrase-level and the sentence-level patterns.
 \begin{figure}[h!]
    \begin{center}
        \captionsetup{justification=centering}
         \includegraphics[width=0.8\linewidth]{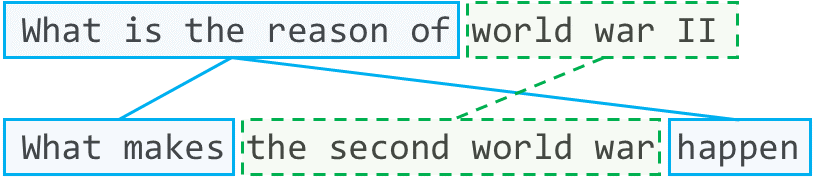}
          \vspace{-3pt}\caption*{\small{\textit{what is the reason of \$x} $\to$ \textit{what makes \$x happen}}\\ \small{\underline{world war II} $\to$ \underline{the second world war}}}
    \end{center}
    \end{figure}

Specifically, the blue part is the sentence-level pattern, which can be expressed as a pair of sentence templates, where \textit{\$x} can be any fragment of text. The green part is the phrase-level pattern, which is a pair of phrases. Table~\ref{tab:examples} shows more examples of paraphrase pairs sampled from WikiAnswers corpus~\footnote{http://knowitall.cs.washington.edu/paralex/} and Quora question pairs~\footnote{https://www.kaggle.com/c/quora-question-pairs}. We can see that the sentence-level paraphrases are more general and abstractive, while the word/phrase-level paraphrases are relatively diverse and domain-specific. Moreover, we notice that in many cases, paraphrasing can be decoupled, i.e., the word-level and phrase-level patterns are mostly independent of the sentence-level paraphrase patterns. 

\begin{table}[t]
    \centering
    \caption{Examples of paraphrase pairs in WikiAnswers and Quora datasets. We manually labeled the sentences with the \textcolor{brandeisblue}{\textit{blue italic words}} being sentence-level and the \textcolor{caribbeangreen}{\underline{green underlined words}} being phrase-level.}\label{tab:examples}
    \vspace{-5pt}
    \resizebox{0.8\linewidth}{!}{
    \begin{tabular}{l}
        \toprule
        \textcolor{brandeisblue}{\textit{What is the population of}}
        \textcolor{caribbeangreen}{\underline{New York}}\textcolor{brandeisblue}{\textit{?}}\\

        \textcolor{brandeisblue}{\textit{How many people is there in}}
        \textcolor{caribbeangreen}{\underline{NYC}}\textcolor{brandeisblue}{\textit{?}}\\
        \midrule

        \textcolor{brandeisblue}{\textit{Who wrote}}
        \textcolor{caribbeangreen}{\underline{the Winnie the Pooh books}}\textcolor{brandeisblue}{\textit{?}}\\

        \textcolor{brandeisblue}{\textit{Who is the author of}}
        \green\underline{winnie the pooh}\textcolor{brandeisblue}{\textit{?}}\\
        \midrule

        \textcolor{brandeisblue}{\textit{What is}}
        \green\underline{the best phone}
        \textcolor{brandeisblue}{\textit{to buy below}}
        \green\underline{15k}\textcolor{brandeisblue}{\textit{?}} \\
        \textcolor{brandeisblue}{\textit{Which are}}
        \textcolor{caribbeangreen}{\underline{best mobile phones}}
        \textcolor{brandeisblue}{\textit{to buy under}}
        \textcolor{caribbeangreen}{\underline{15000}}\textcolor{brandeisblue}{\textit{?}}\\
        \midrule

        \textcolor{brandeisblue}{\textit{How can I be}}
        \textcolor{caribbeangreen}{\underline{a good geologist}}\textcolor{brandeisblue}{\textit{?}}\\
        \textcolor{brandeisblue}{\textit{What should I do to be}}
        \textcolor{caribbeangreen}{\underline{a great geologist}}\textcolor{brandeisblue}{\textit{?}}\\
        \midrule

        \textcolor{brandeisblue}{\textit{How do I}}
        \textcolor{caribbeangreen}{\underline{reword a sentence}}
        \textcolor{brandeisblue}{\textit{to}}
        \textcolor{caribbeangreen}{\underline{avoid plagiarism}}\textcolor{brandeisblue}{\textit{?}}\\

        \textcolor{brandeisblue}{\textit{How can I}}
        \textcolor{caribbeangreen}{\underline{paraphrase my essay}}
        \textcolor{brandeisblue}{\textit{and}}
        \textcolor{caribbeangreen}{\underline{avoid plagiarism}}\textcolor{brandeisblue}{\textit{?}}\\
        \bottomrule
    \end{tabular}
    }
\end{table}


To address this phenomenon in paraphrase generation, we propose \textbf{D}ecomposable \textbf{N}eural \textbf{P}araphrase \textbf{G}enerator (DNPG). Specifically, the DNPG consists of a \textit{separator}, multiple \textit{encoders} and \textit{decoders}, and an \textit{aggregator}. The \textit{separator} first partitions an input sentence into segments belonging to different granularities, which are then processed by multiple granularity-specific encoders and decoders in parallel. Finally the \textit{aggregator} combines the outputs from all the decoders to produce a paraphrase of the input.

We explore three advantages of the DNPG:

\paragraph{Interpretable} In contrast to the existing Seq2Seq models, we show that DNPG can automatically learn the paraphrasing transformation separately at lexical/phrasal and sentential levels. Besides generating a paraphrase given a sentence, it can meanwhile interpret its prediction by extracting the associated paraphrase patterns at different levels, similar to the examples shown above.

\paragraph{Controllable} The model allows the user to control the generation process precisely. By employing DNPG, the user can specify the part of the sentence being fixed while the rest being rephrased at a particular level.

\paragraph{Domain-adaptable} In this work, we assume that high-level paraphrase patterns are more likely to be shared across different domains. With all the levels coupled together, it is difficult for conventional Seq2Seq models to well adapt to a new domain. The DNPG model, however, can conduct paraphrase at abstractive (sentential) level individually, and thus be more capable of performing well in domain adaptation. Concretely, we develop a method for the DNPG to adapt to a new domain with only non-parallel data.

We verify the DNPG model on two large-scale paraphrasing datasets and show that it can generate paraphrases in a more controllable and interpretable way while preserving the quality. Furthermore, experiments on domain adaptation show that DNPG performs significantly better than the state-of-the-art methods.
The technical contribution of this work is of three-fold:
\begin{enumerate}
  \item We propose a novel Seq2Seq model that decomposes the paraphrase generation into learning paraphrase patterns at different granularity levels separately.
  \item We demonstrate that the model achieves more interpretable and controllable generation of paraphrases.
  \item Based on the proposed model, we develop a simple yet effective method for unsupervised domain adaptation.
\end{enumerate}

\section{Decomposable Neural Paraphrase Generator}
This section explains the framework of the proposed DNPG model. We first give an overview of the model design and then elaborate each component in detail.
\subsection{Model Overview}

\begin{figure}[h!]
    \begin{center}
         \includegraphics[width=\linewidth]{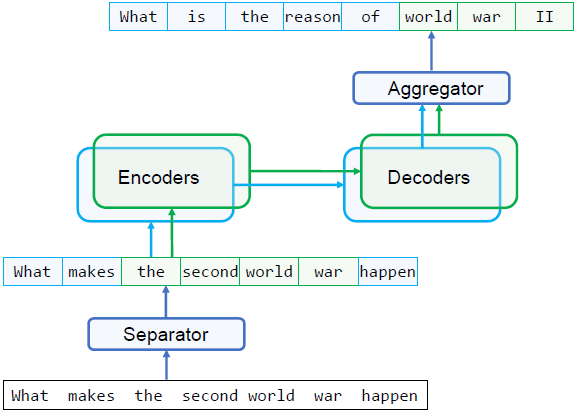}
         \caption{Model Architecture.} \label{fig:model}
    \end{center}
    \end{figure}

As illustrated in Figure~\ref{fig:model}, DNPG consists of four components: a separator, multi-granularity encoders and decoders (denoted as $m$-encoder and $m$-decoder respectively), and an aggregator. The $m$-encoder and $m$-decoder are composed of multiple independent encoders and decoders, with each corresponding to a specific level of granularity. Given an input sentence of words $X=[\rmx_{1}, \ldots, \rmx_{L}]$ with length $L$, the separator first determines the granularity label for each word, denoted as $Z=[z_1,\ldots,z_{L}]$. After that, the input sentence $X$ together with its associated labels $Z$ are fed into $m$-encoder in parallel and summarized as
    \begin{equation}
        \begin{aligned}
        U_{z} = m\text{-encoder}_{z}(X, Z),
        \end{aligned}
    \end{equation}
where the subscript $z$ denotes the granularity level. At the decoding stage, each decoder can individually predict the probability of generating the next word $\rmy_t$ as
    \begin{equation}
      P_{z}(\rmy_{t}|\rmy_{1:t-1}, X) = m\text{-decoder}_{z}(U_z, \rmy_{1:t-1}).
    \end{equation}

Finally, the aggregator combines the outputs of all the decoders and make the final prediction of the next word:
   \begin{equation}\label{final}
        \begin{aligned}
       P(\rmy_{t} &|\rmy_{1:t-1}, X) = \\
        &\sum_{z_t}P_{z_{t}}(\rmy_{t}|\rmy_{1:t-1}, X)P(z_t|\rmy_{1:t-1}, X).
        \end{aligned}
    \end{equation}

Here $P(z_t|\rmy_{1:t-1}, X)$ is computed as the probability of being at the granularity level $z_t$, and $P_{z_t}(\rmy_{t}|\rmy_{1:t-1}, X)$ is given by the decoder $m\text{-decoder}_{z_t}$ at level $z_t$.

The choice of the encoder and decoder modules of DNPG can be quite flexible, for instance long-short term memory networks (LSTM)~\citet{hochreiter1997long} or convolutional neural network (CNN)~\cite{lecun1998gradient}. In this work, the $m$-encoder and $m$-decoder are built based on the Transformer model~\cite{vaswani2017attention}. Besides, we employ LSTM networks to build the separator and aggregator modules. Without loss of generality, we consider two levels of granularity in our experiments, that is, $z=0$ for the lexical/phrasal level and $z=1$ for the sentential level.

\subsection{Separator}
For each word $\rmx_l$ in the sentence, we assign a latent variable $z_l$ indicating its potential granularity level for paraphrasing. This can be simply formulated as a sequence labeling process. In this work we employ the stacked LSTMs to compute the distribution of the latent variables recursively:
    \begin{equation}\label{seq-tag}
      \begin{aligned}
        & h_{l} = \text{BiLSTM}([\rmx_{l};h_{l-1}, h_{l+1}]) \\
        & g_{l} = \text{LSTM}([h_{l}, z_{l-1};g_{l-1}]) \\
        & P(z_{l}|X)=\text{GS}(W_{g} g_{l}, \tau)
      \end{aligned}
    \end{equation}
where $h_l$ and $g_l$ represent the hidden states in the LSTMs and $\text{GS}(\cdot, \tau)$ denotes the Gumbel-Softmax function~\citep{jang2016categorical}. The reason of using Gumbel-Softmax is to make the model differentiable, and meanwhile produce the approximately discrete level for each token. $\tau$ is the temperature controlling the closeness of $z$ towards $0$ or $1$.
\subsection{Multi-granularity encoder and decoder}
We employ the Transformer architecture for the encoders and decoders in DNPG. Specifically, the phrase-level Transformer is composed of $m\text{-encoder}_{0}$ and $m\text{-decoder}_{0}$, which is responsible for capturing the local paraphrasing patterns. The sentence-level Transformer is composed of $m\text{-encoder}_{1}$ and $m\text{-decoder}_{1}$, which aims to learn the high-level paraphrasing transformations. Based on the Transformer design in~\citet{vaswani2017attention}, each encoder or decoder is composed of positional encoding, stacked multi-head attention, layer normalization, and feed-forward neural networks. The multi-head attention in the encoders contains self-attention while the one in the decoders contains both self-attention and context-attention. We refer readers to the original paper for details of each component. In order to better decouple paraphrases at different granularity levels, we introduce three inductive biases to the modules by varying the model capacity and configurations in the positional encoding and multi-head attention modules. We detail them hereafter.

\paragraph{Positional Encoding:}We adopt the same variant of the positional encoding method in~\citet{vaswani2017attention}, that is, the sinusoidal function:
    \begin{equation}\label{pe}
    \begin{aligned}
      &\text{PE}(pos, 2d) = \sin(p/10000^{2d/\mathbb{D}})\\
      &\text{PE}(pos, 2d+1) = \cos(p/10000^{2d/\mathbb{D}})\\
    \end{aligned}
    \end{equation}
For phrase-level Transformer, we use the original position, i.e., $p:=pos$. For the sentence-level Transformer, in order to make the positional encoding insensitive to the lengths of the phrase-level fragment, we set:
    \begin{equation}
      p = \sum_{i=1}^{pos}P(z_{i}=1)
    \end{equation}

\begin{figure}[h!]
    \begin{center}
         \includegraphics[width=0.8\linewidth]{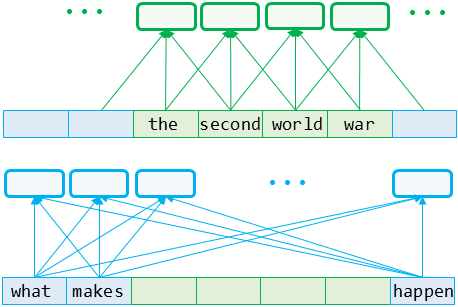}
         \caption{Attention: phrase-level self-attention (upper) and sentence-level self-attention (lower).}\label{fig:encoder} 
    \end{center}
    \end{figure}

\paragraph{Multi-head Attention:}We modify the self-attention mechanism in the encoders and decoders by setting a different receptive field for each granularity, as illustrated in Figure~\ref{fig:encoder}. Specifically, for the phrase-level model, we restrict each position in the encoder and decoder to attend only the adjacent $n$ words ($n = 3$), so as to mainly capture the local composition. As for the sentence-level model, we allow the self-attention to cover the entire sentence, but only those words labeled as sentence-level (i.e., $z_{l}=1$) are visible. In this manner, the model will focus on learning the sentence structures while ignoring the low-level details. To do so, we re-normalize the original attention weights $\alpha_{t,l}$ as
    \begin{equation}\label{renorm-attn}
      \alpha^{'}_{t, l} = \frac{P(z_{l}=1)\alpha_{t,l}}{\sum_{l=1}^L P(z_{l}=1)\alpha_{t,l}}.
    \end{equation}
We also restrict the decoder at $z$ level only access the position $l:z_{l}=z$ at encoder in the same way.

\paragraph{Model Capacity:}We choose a larger capacity for the phrase-level Transformer over the sentence-level Transformer. The intuition behind is that lexical/phrasal paraphrases generally contain more long-tail expressions than the sentential ones. In addition, the phrase-level Transformer is equipped with the copying mechanism~\citep{gu2016incorporating}. Thus, the probability of generating the target word $y_{t}$ by the $m\text{-decoder}_{0}$ is:
    \begin{equation}
    \begin{aligned}
    P_{z=0}(\rmy_{t}|\rmy_{1:t-1}, X) = &(1-\rho_{t})P_{\text{gen}}(\rmy_{t}|\rmy_{1:t-1}, X)\\
    & +\rho_{t}P_{\text{copy}}(\rmy_{t}|\rmy_{1:t-1}, X)
    \end{aligned}
    \end{equation}
where $\rho_{t}$ is the copying probability, which is jointly learned with the model. Table~\ref{tab:model-spec} summarizes the specifications of the Transformer models for each granularity.

\begin{table}[h!]
    \centering
    \caption{Model Specifications.}\label{tab:model-spec}
    \resizebox{1.0\linewidth}{!}{
    \begin{tabular}{lcc}
        \toprule
        & Phrase-level model & Sentence-level model\\
        \midrule
        Receptive field & Local & Global \\
        Word Visibility & $\{x_l\}_{l=1}^L$ & $\{\rmx_l\}_{l:z_l=1}$ \\
        \#Dimension & 300 & 150 \\
        \#Heads & 6 & 3 \\
        Copy mechanism & Yes & No \\
        \bottomrule
    \end{tabular}
    }
\end{table}

\subsection{Aggregator}
Each Transformer model works independently until generating the final paraphrases.
     \begin{figure}[h!]
    \begin{center}
         \includegraphics[width=0.9\linewidth]{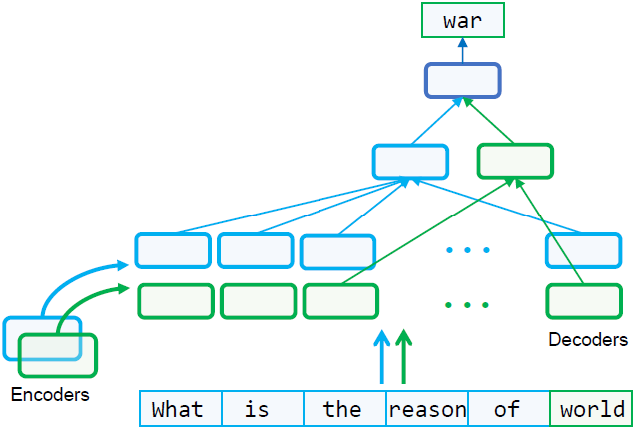}
         \caption{Aggregator.} \label{fig:decoder}
    \end{center}
    \end{figure}
The prediction of the token at $t$-th position is determined by the aggregator, which combines the outputs from the $m$-decoders. More precisely, the aggregator first decides the probability of the next word being at each granularity. The previous word $\rmy_{t-1}$ and the context vectors $c_0$ and $c_1$ given by $m\text{-decoder}_{0}$ and $m\text{-decoder}_{1}$, are fed into a LSTM to make the prediction:
   \begin{equation}
        \begin{aligned}
        &v_t = \text{LSTM}([W_{c}[c_0; c_1; \rmy_{t-1}]; v_{t-1}])\\
        &P(z_t|\rmy_{1:t-1}, X) = \text{GS}(W_v v_t, \tau),
        \end{aligned}
    \end{equation}
where $v_t$ is the hidden state of the LSTM. Then, jointly with the probabilities computed by $m$-decoders, we can make the final prediction of the next word via Eq \eqref{final}.

\subsection{Learning of Separator and Aggregator}
The proposed model can be trained end-to-end by maximizing the conditional probability~\eqref{final}. However, learning from scratch may not be informative for the separator and aggregator to disentangle the paraphrase patterns in an optimal way. Thus we induce weak supervision to guide the training of the model. We construct the supervision based on a heuristic that long-tail expressions contain more rare words. To this end, we first use the word alignment model~\citep{och03:asc} to establish the links between the words in the sentence pairs from the paraphrase corpus. Then we assign the label $z^*=0$ (phrase-level) to $n$ (randomly sampled from $\{1, 2, 3\}$) pairs of aligned phrases that contain most rare words. The rest of the words are labeled as $z^*=1$ (sentence-level).

We train the model with explicit supervision at the beginning, with the following loss function:
    \begin{equation}\label{constrnt}
    \begin{aligned}
      &\mathcal{L} = \sum_{t=1}^T\log  P(\rmy_{t}|\rmy_{1:t-1}, X) + \\
      &\lambda (\sum_{l=1}^L\log P(z^*_{l}|X) +\sum_{t=1}^{T}\log P(z^*_{t}|\rmy_{1:t-1}, X) )
      \end{aligned}
    \end{equation}
where $\lambda$ is the hyper-parameter controlling the weight of the explicit supervision. In experiments, we decrease $\lambda $ gradually from 1 to nearly 0.


\section{Applications and Experimental Results}
We verify the proposed DNPG model for paraphrase generation in three aspects: interpretability, controllability and domain adaptability. We conduct experiments on WikiAnswers paraphrase corpus~\citep{Fader14} and Quora duplicate question pairs, both of which are questions data. While the Quora dataset is labeled by human annotators, the WikiAnswers corpus is collected in a automatic way, and hence it is much noisier. There are more than 2 million pairs of sentences on WikiAnswers corpus. To make the application setting more similar to real-world applications, and more challenging for domain adaptation, we use a randomly sampled subset for training. The detailed statistics are shown in Table \ref{tab:data-stat}.

\begin{table}[h!]
    \centering
    \caption{Statistics of the paraphrase datasets.}\label{tab:data-stat}
    \vspace{-3pt}
    \resizebox{0.6\linewidth}{!}{
    \begin{tabular}{lcc}
        \toprule
        & WikiAnswers & Quora \\
        \midrule
        Training & 500K & 100K \\
        Validation & 6K & 4K \\
        Test & 20K & 20K \\
        \bottomrule
    \end{tabular}
    }
\end{table}

\subsection{Implementation and Training Details} As the words in the WikiAnswers are all stemmed and lower case, we do the same pre-processing on Quora dataset. For both datasets, we truncate all the sentences longer than 20 words. For the models with copy mechanism, we maintain a vocabulary of size 8K. For the other baseline models besides vanilla Transformer, we include all the words in the training sets into vocabulary to ensure that the improvement of our models does not come from solving the out-of-vocabulary issue. For a fair comparison, we use the Transformer model with similar number of parameters with our model. Specifically, it is with $3$ layers, model size of 450 dimensions, and attention with 9 heads. We use early stopping to prevent the problem of over-fitting. We train the DNPG with Adam optimizer~\cite{kingma2014adam}. We set the learning rate as $5e-4$, $\tau$ as 1 and $\lambda$ as 1 at first, and then decrease them to $1e-4$, $0.9$ and $1e-2$ after 3 epochs. We set the hyper-parameters of models and optimization in all other baseline models to remain the same in their original works. We implement our model with PyTorch~\cite{paszke2017automatic}.

\subsection{Interpretable Paraphrase Generation}\label{int-pps}

\begin{table*}[t]
    \centering
    \parbox[t]{0.9\textwidth}{
    \centering
    \caption{In-domain performance of paraphrase generation.}\label{tab:performance}
    \resizebox{0.9\linewidth}{!}{
    \begin{tabular}{lccccccccc}
        \toprule
        & \multicolumn{4}{c}{Quora} && \multicolumn{4}{c}{WikiAnswers}\\
        \cmidrule(lr){2-5}\cmidrule(lr){7-10}
        Models & BLEU & iBLEU & ROUGE-1 & ROUGE-2 && BLEU & iBLEU & ROUGE-1 & ROUGE-2\\
        \midrule
        ResidualLSTM & 17.57 & 12.67 & 59.22 & 32.40 && 27.36 & 22.94 & 48.52 & 18.71 \\
        VAE-SVG-eq & 20.04 & 15.17 & 59.98 & 33.30 && 32.98 & 26.35 & 50.93 & 19.11 \\
        Pointer-generator & 22.65 & 16.79 & 61.96 & 36.07 && 39.36 & 31.98 & 57.19 & 25.38\\
        Transformer & 21.73 & 16.25 & 60.25 & 33.45 && 33.01 & 27.70 & 51.85 & 20.70  \\ 
        Transformer+Copy & 24.77 & 17.98 & 63.34 & 37.31 && 37.88 & 31.43 & 55.88 & 23.37 \\ 
        \midrule
        DNPG (ours) & \textbf{25.03} & \textbf{18.01} & \textbf{63.73} & \textbf{37.75} && \textbf{41.64} & \textbf{34.15} & \textbf{57.32} & \textbf{25.88}\\
        \bottomrule
    \end{tabular}
    }
    \vspace{10pt}
    }
    \hfill
    \parbox[t]{0.9\textwidth}{
    \centering
    \caption{Performance of paraphrase generation on domain adaptation (source $\to$ target).}\label{tab:adap-performance}
    \resizebox{0.9\linewidth}{!}{
    \begin{tabular}{lccccccccc}
        \toprule
        & \multicolumn{4}{c}{WikiAnswers$\to$ Quora} && \multicolumn{4}{c}{Quora$\to$WikiAnswers}\\
        \cmidrule(lr){2-5}\cmidrule(lr){7-10}
        Models & BLEU & iBLEU & ROUGE-1 & ROUGE-2 && BLEU & iBLEU & ROUGE-1 & ROUGE-2\\
        \midrule
        Pointer-generator & 6.96 & 5.04 & 41.89& 12.77 && 27.94 & 21.87 & 53.99 & 20.85 \\
        Transformer+Copy & 8.15 & 6.17 & 44.89 & 14.79  &&  29.22 & 23.25 & 53.33 & 21.02\\
        DNPG (ours) & 10.00 & 7.38 & 47.53 & 18.89 && 31.84 & 24.22 & 54.87 & 22.27 \\
       \midrule
        Shallow fusion & 7.95 & 6.04 & 44.87 & 14.79 && 29.76 & 22.57 & 53.54 & 20.68\\
        MTL & 6.37 & 4.90 & 37.64 & 11.83 && 23.65 & 18.34 & 48.19 & 17.53\\
        MTL+Copy & 9.83 & 7.22 & 47.08 & 19.03 && 30.78 & 21.87 & 54.10 & 21.08\\
        Adapted DNPG (ours) & \textbf{16.98} & \textbf{10.39} & \textbf{56.01} & \textbf{28.61} && \textbf{35.12} & \textbf{25.60} & \textbf{56.17} & \textbf{23.65}\\
        \bottomrule
    \end{tabular}
    }
    }
\end{table*}

First, we evaluate our model quantitatively in terms of automatic metrics such as BLEU~\citep{papineni2002bleu}, ROUGE~\citep{lin2004rouge}, which have been widely used in previous works on paraphrase generation. In addition, we include iBLEU \citep{sun2012joint}, which penalizes repeating the source sentence in its paraphrase. We use the same hyper-parameter in their original work. We compare DNPG with four existing neural-based models: ResidualLSTM~\cite{prakash2016neural}, VAE-SVG-eq~\citep{gupta2017deep}, pointer-generator~\citep{see2017get} and the Transformer~\citep{vaswani2017attention}, the latter two of which have been reported as the state-of-the-art models in~\citet{li2017paraphrase} and~\citet{wang2018task} respectively. For a fair comparison, we also include a Transformer model with copy mechanism. Table~\ref{tab:performance} shows the performances of the models, indicating that DNPG achieves competitive performance in terms of all the automatic metrics among all the models. In particular, the DNPG has similar performance with the vanilla Transformer model on Quora dataset, while significantly performs better on WikiAnswers. The reason maybe that the DNPG is more robust to the noise, since it can process the paraphrase in an abstractive way. It also validates our assumption that paraphrasing can be decomposed in terms of granularity. When the training data of high quality is available, the transformer-based models significantly outperforms the LSTM-based models.


Besides the quantitative performance, we demonstrate the interpretability of DNPG. Given an input sentence, the model can not only generate its paraphrase but also predict the granularity level of each word. By using the predicted granularity levels and the context attentions in the Transformer, we are able to extract the phrasal and sentential paraphrase patterns from the model. Specifically, we extract the sentential templates $\bar{X}$ of $X$ (or $\bar{Y}$ of $Y$) by substituting each fragment of words at the phrasal level by a unique placeholder such as \textit{\$x}. The extraction process is denoted as $\bar{X}=\mathcal{T}(X,Z)=[\bar{\rmx}_{1},\ldots,\bar{\rmx}_{\bar{L}}]$, where the element $\bar{\rmx}_{\bar{l}}$ is either a placeholder or a word labeled as sentence-level. Through the attention weights, we ensure that the pair of aligned fragment share the same placeholder in $\{\bar{X}, \bar{Y}\}$. The whole generation and alignment process is detailed in Appendix A. Each pair of fragments sharing the same placeholder are extracted as the phrasal paraphrase patterns.

Table~\ref{tab:gen-int} gives examples of the generated paraphrases and the corresponding extracted templates. For instance, the model learns a sentential paraphrasing pattern: $\bar{X}$: \textit{what is \$x's \$y} $\to$ $\bar{Y}$: \textit{what is the \$y of \$x}, which is a common rewriting pattern applicable in general practice. The results clearly demonstrate the ability of DNPG to decompose the patterns at different levels, making its behaviors more interpretable.

\begin{table*}[t]
    \centering
    \caption{Examples of the generated paraphrases and extracted patterns at each granularity level by DNPG.}\label{tab:gen-int}
    \resizebox{1.0\linewidth}{!}{
    \begin{tabular}{p{5.5cm}p{5.5cm}p{5cm}p{5cm}}
        \toprule
        Input Sentence& Generate Paraphrase & Sentential Paraphrase Patterns & Phrasal Paraphrase Patterns \\
        \midrule
        what is the technique for prevent suicide? & how can you prevent suicide? & \textit{what is the technique for \$x}\newline$\to$ \textit{how can you \$x} & - \\
        what is the second easiest island? & what is the 2nd easiest island? & - & \underline{second easiest island}\newline$\to$ \underline{2nd easiest island}\\
        what is rihanna brother's name? & what is the name of rihanna's brother? & \textit{what is \$x's \$y}\newline$\to$ \textit{what is the \$y of \$x} & \underline{rihanna brother}\newline$\to$ \underline{rihanna's brother}\\
       do anyone see the relation between greek god and hindu god? & what is the relationship between the greek god and hindu god? & \textit{do anyone see the \$x between \$y} $\to$  \textit{what is the \$x between the \$y}& \underline{relation} $\to$ \underline{relationship} \\
        \bottomrule
    \end{tabular}
    }
\end{table*}

\subsection{Controllable Paraphrase Generation}\label{con-pps}
The design of the DNPG model allows the user to control the generating process more precisely. Thanks to the decomposable mechanisms, it is flexible for the model to conduct either sentential paraphrasing or phrasal paraphrasing individually. Furthermore, instead of using the learned separator, the user can manually specify the granularity labels of the input sentence and then choose the following paraphrasing strategies.

\paragraph{Sentential paraphrasing} is performed by restricting the phrase-level decoder ($m\text{-decoder}_{0}$) to copying from the input at the decoding stage, i.e., keeping the copy probability $\rho_t=1$. To ensure that the phrasal parts are well preserved, we replace each phrase-level fragment by a unique placeholder and recover it after decoding.

\paragraph{Phrasal paraphrasing} is performed with sentence template being fixed. For each phrase-level fragment, paraphrase is generated by $m\text{-decoder}_{0}$ only and the generation stopped at $t: z_{t}=1$.

Once the beam search of size $B$ finished, there are $B$ paraphrase candidates $\hat{Y}_{b}$. We pick up the one with the best accuracy and readability. Specifically, we re-rank them by $P(\hat{Y}_{b}|X,Z)$ calculated by the full model of DNPG.

Given a sentence, we manually label different segments of words as phrase-level, and employ the model to conduct sentential and phrasal paraphrasing individually. With the manual labels, the model automatically selects different paraphrase patterns for generation. Table \ref{tab:gen-con} shows examples of the generated results by different paraphrasing strategies. As demonstrated by the examples, DNPG is flexible enough to generate paraphrase given different sentence templates and phrases.

Controllable generation is useful in downstream applications, for instance, data augmentation in the task-oriented dialogue system. Suppose we have the user utterance \textit{book a flight from} \underline{New York} \textit{to} \underline{London} and want to produce more utterances with the same intent. With the DNPG, we can conduct sentential paraphrasing and keep the slot values fixed, e.g. \textit{buy an airline ticket to} \underline{London} \textit{from} \underline{New York}.


\begin{table}[t!]
    \centering
    \caption{Examples of controllable generation of paraphrase. The words with underline are labeled as phrase-level and the ones in \textit{italic} form are at sentence-level. The strategy \textit{All} is referred as the fully automatic generation.}\label{tab:gen-con}
    \vspace{-3pt}
    \resizebox{\linewidth}{!}{
    \begin{tabular}{p{6.2cm}p{1.2cm}p{5cm}}
        \toprule
        Input sentence \& labels & Strategy & Generated Paraphrase \\
        \midrule
        what is the value of a 1961 us cent? & All & what is the 1961 nickel 's value?\\
        \textcolor{brandeisblue}{\textit{what is the}}
        \textcolor{caribbeangreen}{\underline{value}}
        \textcolor{brandeisblue}{\textit{of}}
        \textcolor{caribbeangreen}{\underline{a 1961 us cent}}\textcolor{brandeisblue}{\textit{?}}

        & Phrase

        & \textcolor{brandeisblue}{\textit{what is the}}
        \textcolor{caribbeangreen}{\underline{price}}
        \textcolor{brandeisblue}{\textit{of}}
        \textcolor{caribbeangreen}{\underline{a 1961 nickel}}\textcolor{brandeisblue}{\textit{?}}\\
        \textcolor{brandeisblue}{\textit{what is the}}
        \textcolor{caribbeangreen}{\underline{value}}
        \textcolor{brandeisblue}{\textit{of}}
        \textcolor{caribbeangreen}{\underline{a 1961 us cent}}\textcolor{brandeisblue}{\textit{?}}

        & Sentence

        & \textcolor{brandeisblue}{\textit{what is the}}
        \textcolor{caribbeangreen}{\underline{1961 us cent}}
        \textcolor{brandeisblue}{\textit{'s}}
        \textcolor{caribbeangreen}{\underline{value}}\textcolor{brandeisblue}{\textit{?}}\\
        \textcolor{brandeisblue}{\textit{what is the value of}}
        \textcolor{caribbeangreen}{\underline{a 1961 us cent}}\textcolor{brandeisblue}{\textit{?}}

        & Phrase

        & \textcolor{brandeisblue}{\textit{what is the value of}}
        \textcolor{caribbeangreen}{\underline{a 1961 nickel}}\textcolor{brandeisblue}{\textit{?}}\\

        \textcolor{brandeisblue}{\textit{what is the value of}}
        \textcolor{caribbeangreen}{\underline{a 1961 us cent}}\textcolor{brandeisblue}{\textit{?}}

        & Sentence

        & \textcolor{brandeisblue}{\textit{how much is}}
        \textcolor{caribbeangreen}{\underline{a 1961 us cent}} \textcolor{brandeisblue}{\textit{worth}}\textcolor{brandeisblue}{\textit{?}}\\
        \midrule

        what should i do to avoid sleep in class? & All & how do i prevent sleep in class?\\

        \textcolor{brandeisblue}{\textit{what should i do to}} \textcolor{caribbeangreen}{\underline{avoid sleep in class}}\textcolor{brandeisblue}{\textit{?}}

        & Phrase

        & \textcolor{brandeisblue}{\textit{what should i do to}} \textcolor{caribbeangreen}{\underline{prevent sleep in class}}\textcolor{brandeisblue}{\textit{?}}\\

        \textcolor{brandeisblue}{\textit{what should i do to}} \textcolor{caribbeangreen}{\underline{avoid sleep in class}}\textcolor{brandeisblue}{\textit{?}}

        & Sentence

        & \textcolor{brandeisblue}{\textit{how do i}}
        \textcolor{caribbeangreen}{\underline{avoid sleep in class}}\textcolor{brandeisblue}{\textit{?}}\\

        \textcolor{brandeisblue}{\textit{what should i do to avoid}} \textcolor{caribbeangreen}{\underline{sleep in class}}\textcolor{brandeisblue}{\textit{?}}
        & Phrase

        & \textcolor{brandeisblue}{\textit{what should i do to avoid}} \textcolor{caribbeangreen}{\underline{fall sleep during class}}\textcolor{brandeisblue}{\textit{?}} \\
        \textcolor{brandeisblue}{\textit{what should i do to avoid}} \textcolor{caribbeangreen}{\underline{sleep in class}}\textcolor{brandeisblue}{\textit{?}}

        & Sentence

        & \textcolor{brandeisblue}{\textit{what should i do if i don't want to}}
        \textcolor{caribbeangreen}{\underline{sleep in class}}\textcolor{brandeisblue}{\textit{?}} \\
        \bottomrule
    \end{tabular}
   }
\end{table}

\subsection{Unsupervised Domain Adaptation}
\begin{figure}[h!]
    \begin{center}
         \includegraphics[width=\linewidth]{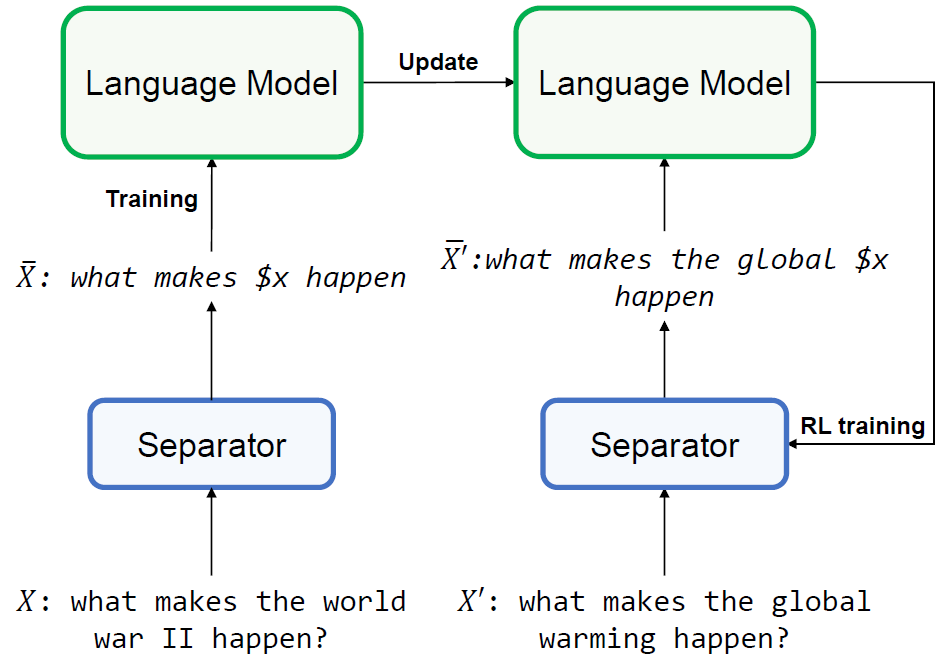}
         \caption{Left: Training of language model in the source domain; Right: RL training of separator in the target domain.} \label{fig:rl-adap}
    \end{center}\vspace{-10pt}
    \end{figure}

\vspace{-3pt}
Existing studies on paraphrase generation mainly focus on the in-domain setting with a large-scale parallel corpus for training. In practice, there is always a need to apply the model in a new domain, where no parallel data is available. We formulate it as an unsupervised domain adaptation problem for paraphrase generation. Based on the observation that the sentence templates generated by DNPG tend to be more general and domain-insensitive, we consider directly performing the sentential paraphrase in the target domain as a solution. However, the language model of the source and target domains may differ, we therefore fine-tune the separator of DNPG so that it can identify the granularity of the sentence in the target domain more accurately. Specifically, to adapt the separator $P_{\textrm{sep}}(Z|X)$ to the target domain, we employ a reinforcement learning (RL) approach by maximizing the accumulative reward:
\begin{equation}\label{fig:domain-adap}
    \mathcal{R}_{\text{separator}} = \mathbb{E}_{P_{\textrm{sep}}(Z|X)}\sum_{l=1}^{L}r_{l}(z_{1:l}, X).
\end{equation}
We define the reward functions based on the principle that the source and target domain share the similar sentence templates. We first train a neural language model, specifically LSTM, with the sentence templates in the source domain, with the conditional probability denoted as $P_{\text{LM}}(\bar{\rmx}_{\bar{l}}|\bar{\rmx}_{1:\bar{l}-1})$. In the target domain, the template language model is employed as a reward function for separator. Formally we define the reward $r_l$ at position $l$ as:
\begin{equation}\label{reward}
    \begin{aligned}
    r_{l}(z_{1:l}, X) &= \alpha P_{\textrm{LM}}(\bar{\rmx}_{\bar{l}}|\bar{\rmx}_{1:\bar{l}-1}),
    \end{aligned}
\end{equation}
where the template $\bar{\rmx}_{1:\bar{l}} = \mathcal{T}(X, z_{1:l})$ is extracted in the way as detailed in Section \ref{int-pps}. And $\alpha$ is a scaling factor that penalizes the long fragment labeled as phrase-level, since more informative sentence templates are preferred. With the reward, the separator is further tuned with the policy gradient method~\cite{williams1992simple,sutton2000policy}. To bridge the gap between training and testing of the Transformer models in different domain, we fine-tune the DNPG model on the sentential paraphrase patterns extracted in source domain. Since only the unlabeled data in the target domain is needed to fine-tune separator, the domain adaptation can be done incrementally. An overview of the complete training process is illustrated in Figure \ref{fig:rl-adap}. We refer the model fine-tuned in this way as Adapted DNPG.

We evaluate the performance of the original DNPG and the Adapted DNPG in two settings of domain transfer: 1) Quora dataset as the source domain and WikiAnswers as the target domain, denoted as Quora$\to$WikiAnswers, and 2) in reverse as WikiAnswers$\to$Quora. For the baseline models, in addition to the pointer-generator network and the Transformer model with copy mechanism (denoted as Transformer+Copy), we use the shallow fusion~\cite{gulcehre2015using} and the multi-task learning (MTL)~\cite{domhan2017using} that harness the non-parallel data in the target domain for adaptation. For fair comparisons, we use the Transformer+Copy as the base model for shallow fusion and implement a variant of MTL with copy mechanism (denoted as MTL+Copy). Table \ref{tab:adap-performance} shows performances of the models in two settings. DNPG achieves better performance over the pointer-generator and Transformer-based model, and has the competitive performance with MTL+Copy, which accesses target domain for training. With a fine-tuned separator, Adapted DNPG outperforms other models significantly on Quora$\to$WikiAnswers. When it comes to WikiAnswers$\to$Quora, where domain adaptation is more challenging since the source domain is noisy, the margin is much larger. The main reason is that the original meaning can be preserved well when the paraphrasing is conducted at the sentential level only. For an intuitive illustration, We show examples of the generated paraphrases from Adapted DNPG and MTL+Copy in Table \ref{tab:gen-adap} in Appendix. It is shown that the sentential paraphrasing is an efficient way to reuse the general paraphrase patterns and meanwhile avoid mistakes on rephrasing domain-specific phrases. However, it is at the expense of the diversity of the generated paraphrases. We leave this problem for future work.

To further verify the improvement of Adapted DNPG, we conduct the human evaluation on the WikiAnswers$\to$Quora setting. We have six human assessors to evaluate 120 groups of paraphrase candidates given the input sentences. Each group consists of the output paraphrases from MTL+Copy, DNPG and Adapted DNPG as well as the reference. The evaluators are asked to rank the candidates from $1$ (best) to $4$ (worst) by their readability, accuracy and surface dissimilarity to the input sentence. The detailed evaluation guide can be found in Appendix B. Table \ref{tab:human-eval} shows the mean rank and inter-annotator agreement (Cohen's kappa) of each model. Adapted DNPG again significantly outperforms MTL+Copy by a large margin ($p$-value $<0.01$). The performance of the original DNPG and MTL+Copy has no significant difference ($p$-value $=0.18$). All of the inter-annotator agreement is regarded as \textit{fair} or above.
\begin{table}
  \centering
  \caption{Human Evaluation in WikiAnswers$\to$Quora}\label{tab:human-eval}
  \resizebox{0.8\linewidth}{!}{
  \begin{tabular}{lccc}
    \toprule
    Models & Mean Rank & Agreement\\
    \midrule
    MTL+Copy & 3.22 & 0.446\\
    \midrule
    Naive DNPG & 3.13 & 0.323\\
    Adapted DNPG & \textbf{1.79} & 0.383\\
    \midrule
    Reference & 1.48 & 0.338\\
    \bottomrule
  \end{tabular}
  }
\end{table}

\vspace{-10pt}
\subsection{Ablation Studies and Discussion}
We quantify the performance gain of each inductive bias we incorporated in the DNPG model. Specifically, we compare the DNPG with three variants: one with vanilla attention modules, one with vanilla positional encoding and the one uses vanilla softmax. We train them with the training set of WikiAnswers and test in the validation set of Quora. The results are shown in Table \ref{tab:abl}, which shows that each inductive bias has a positive contribution. It further proves that the decomposition mechanism allows the model to capture more abstractive and domain-invariant patterns. We also note that there is a large drop without the constraints on multi-head attention, which is a core part of the decomposition mechanism. We investigate the effect of the weak supervision for separator and aggregator by setting $\lambda$ as 0. Though there is not a significant drop on quantitative performance, we observe that the model struggles to extract meaningful paraphrase patterns. It means that explicit supervision for separator and aggregator can make a difference and it does not need to be optimal. It opens a door to incorporate symbolic knowledge, such as regular expression of sentence templates, human written paraphrase patterns, and phrase dictionary, into the neural network. Through training in a parallel corpus, DNPG can generalize the symbolic rules.

\begin{table}
  \centering
  \caption{Ablation Study in WikiAnswers$\to$Quora}\label{tab:abl}
  \resizebox{0.85\linewidth}{!}{
  \begin{tabular}{lccc}
    \toprule
    Model Variants & BLEU & iBLEU \\
    \midrule
    DNPG & 9.84 & 7.40\\
    \midrule
    w/ Vanilla Multi-Head Attention & 7.65 & 5.86 \\
    w/ Vanilla Positional Encoding & 9.46 & 7.08 \\
    w/ Vanilla Softmax & 9.30 & 7.01 \\
    \bottomrule
  \end{tabular}
  }
\end{table}

\section{Related Work}
\subsection{Neural Paraphrase Generation}
Most of the existing neural methods of paraphrase generation focus on improving the in-domain quality of generated paraphrases. \citet{prakash2016neural} and \citet{ma2018query} adjust the network architecture for larger capacity. \citet{cao2017joint} and~\citet{wang2018task} utilize external resources, in other words, phrase dictionary and semantic annotations. \citet{li2017paraphrase} reinforce the paraphrase generator by a learnt reward function. Although achieving state-of-the-art performances, none of the above work considers the paraphrase patterns at different levels of granularity. Moreover, their models can generate the paraphrase in a neither interpretable nor a fine-grained controllable way. In~\citet{iyyer2018adversarial}'s work, the model is trained to produce a paraphrase of the sentence with a given syntax. In this work, we consider automatically learning controllable and interpretable paraphrasing operations from the corpus. This is also the first work to consider scalable unsupervised domain adaptation for neural paraphrase generation.

\subsection{Controllable and Interpretable Text Generation}
There is extensive attention on controllable neural sequence generation and its interpretation. A line of research is based on variational auto-encoder (VAE), which captures the implicit~\cite{gupta2017deep, li2017deep} or explicit information~\citep{hu2017toward, liao2018quase} via latent representations. Another approach is to integrate probabilistic graphical model, e.g., hidden semi-Markov model (HSMM) into neural network~\citep{wiseman2018learning, dai2016recurrent}. In these works, neural templates are learned as a sequential composition of segments controlled by the latent states, and be used for language modeling and data-to-text generation. Unfortunately, it is non-trivial to adapt this approach to the Seq2Seq learning framework to extract templates from both the source and the target sequence.

\subsection{Domain Adaptation for Seq2Seq Learning}
Domain adaptation for neural paraphrase generation is under-explored. To our best knowledge, \citet{su2017cross}'s work is the only one on this topic. They utilize the pre-trained word embedding and include all the words in both domains to vocabulary, which is tough to scale. Meanwhile, we notice that there is a considerable amount of work on domain adaptation for neural machine translation, another classic sequence-to-sequence learning task. However, most of them require parallel data in the target domain~\cite{wang2017sentence, wang2017instance}. In this work, we consider unsupervised domain adaptation, which is more challenging, and there are only two works that are applicable. \citet{gulcehre2015using} use the language model trained in the target domain to guide the beam search. \citet{domhan2017using} optimize two stacked decoders jointly by learning language model in the target domain and learning to translate in the source domain. In this work, we utilize the similarity of sentence templates in the source and target domains. Thanks to the decomposition of multi-grained paraphrasing patterns, DNPG can fast adapt to a new domain without any parallel data.

\section{Conclusion}
In this paper, we have proposed a neural paraphrase generation model, which is equipped with a decomposition mechanism. We construct such mechanisms by latent variables associated with each word, and a couple of Transformer models with various inductive biases to focus on paraphrase patterns at different levels of granularity. We further propose a fast and incremental method for unsupervised domain adaptation. The quantitative experiment results show that our model has competitive in-domain performance compared to the state-of-the-art models, and outperforms significantly upon other baselines in domain adaptation. The qualitative experiments demonstrate that the generation of our model is interpretable and controllable. In the future, we plan to investigate more efficient methods of unsupervised domain adaptation with decomposition mechanism on other NLP tasks.


\bibliography{reference}

\begin{thebibliography}{31}
\expandafter\ifx\csname natexlab\endcsname\relax\def\natexlab#1{#1}\fi

\bibitem[{Cao et~al.(2017)Cao, Luo, Li, and Li}]{cao2017joint}
Ziqiang Cao, Chuwei Luo, Wenjie Li, and Sujian Li. 2017.
\newblock Joint copying and restricted generation for paraphrase.
\newblock In \emph{Thirty-First AAAI Conference on Artificial Intelligence}.

\bibitem[{Dai et~al.(2016)Dai, Dai, Zhang, Li, and Song}]{dai2016recurrent}
Hanjun Dai, Bo~Dai, Yan-Ming Zhang, Shuang Li, and Le~Song. 2016.
\newblock Recurrent hidden semi-markov model.
\newblock In \emph{International Conference on Learning Representations}.

\bibitem[{Domhan and Hieber(2017)}]{domhan2017using}
Tobias Domhan and Felix Hieber. 2017.
\newblock Using target-side monolingual data for neural machine translation
  through multi-task learning.
\newblock In \emph{Proceedings of the 2017 Conference on Empirical Methods in
  Natural Language Processing}, pages 1500--1505.

\bibitem[{Gu et~al.(2016)Gu, Lu, Li, and Li}]{gu2016incorporating}
Jiatao Gu, Zhengdong Lu, Hang Li, and Victor~OK Li. 2016.
\newblock Incorporating copying mechanism in sequence-to-sequence learning.
\newblock In \emph{Proceedings of the 54th Annual Meeting of the Association
  for Computational Linguistics (Volume 1: Long Papers)}, volume~1, pages
  1631--1640.

\bibitem[{Gulcehre et~al.(2015)Gulcehre, Firat, Xu, Cho, Barrault, Lin,
  Bougares, Schwenk, and Bengio}]{gulcehre2015using}
Caglar Gulcehre, Orhan Firat, Kelvin Xu, Kyunghyun Cho, Loic Barrault, Huei-Chi
  Lin, Fethi Bougares, Holger Schwenk, and Yoshua Bengio. 2015.
\newblock On using monolingual corpora in neural machine translation.
\newblock \emph{arXiv preprint arXiv:1503.03535}.

\bibitem[{Gupta et~al.(2017)Gupta, Agarwal, Singh, and Rai}]{gupta2017deep}
Ankush Gupta, Arvind Agarwal, Prawaan Singh, and Piyush Rai. 2017.
\newblock A deep generative framework for paraphrase generation.
\newblock \emph{arXiv preprint arXiv:1709.05074}.

\bibitem[{Hochreiter and Schmidhuber(1997)}]{hochreiter1997long}
Sepp Hochreiter and J{\"u}rgen Schmidhuber. 1997.
\newblock Long short-term memory.
\newblock \emph{Neural computation}, 9(8):1735--1780.

\bibitem[{Hu et~al.(2017)Hu, Yang, Liang, Salakhutdinov, and
  Xing}]{hu2017toward}
Zhiting Hu, Zichao Yang, Xiaodan Liang, Ruslan Salakhutdinov, and Eric~P Xing.
  2017.
\newblock Toward controlled generation of text.
\newblock \emph{arXiv preprint arXiv:1703.00955}.

\bibitem[{Iyyer et~al.(2018)Iyyer, Wieting, Gimpel, and
  Zettlemoyer}]{iyyer2018adversarial}
Mohit Iyyer, John Wieting, Kevin Gimpel, and Luke Zettlemoyer. 2018.
\newblock Adversarial example generation with syntactically controlled
  paraphrase networks.
\newblock \emph{arXiv preprint arXiv:1804.06059}.

\bibitem[{Jang et~al.(2016)Jang, Gu, and Poole}]{jang2016categorical}
Eric Jang, Shixiang Gu, and Ben Poole. 2016.
\newblock Categorical reparameterization with gumbel-softmax.
\newblock \emph{arXiv preprint arXiv:1611.01144}.

\bibitem[{Kingma and Ba(2014)}]{kingma2014adam}
Diederik~P Kingma and Jimmy Ba. 2014.
\newblock Adam: A method for stochastic optimization.
\newblock \emph{arXiv preprint arXiv:1412.6980}.

\bibitem[{LeCun et~al.(1998)LeCun, Bottou, Bengio, Haffner
  et~al.}]{lecun1998gradient}
Yann LeCun, L{\'e}on Bottou, Yoshua Bengio, Patrick Haffner, et~al. 1998.
\newblock Gradient-based learning applied to document recognition.
\newblock \emph{Proceedings of the IEEE}, 86(11):2278--2324.

\bibitem[{Li et~al.(2017)Li, Lam, Bing, and Wang}]{li2017deep}
Piji Li, Wai Lam, Lidong Bing, and Zihao Wang. 2017.
\newblock Deep recurrent generative decoder for abstractive text summarization.
\newblock \emph{arXiv preprint arXiv:1708.00625}.

\bibitem[{Li et~al.(2018)Li, Jiang, Shang, and Li}]{li2017paraphrase}
Zichao Li, Xin Jiang, Lifeng Shang, and Hang Li. 2018.
\newblock Paraphrase generation with deep reinforcement learning.
\newblock In \emph{Proceedings of the 2018 Conference on Empirical Methods in
  Natural Language Processing}, pages 3865--3878.

\bibitem[{Liao et~al.(2018)Liao, Bing, Li, Shi, Lam, and Zhang}]{liao2018quase}
Yi~Liao, Lidong Bing, Piji Li, Shuming Shi, Wai Lam, and Tong Zhang. 2018.
\newblock Quase: Sequence editing under quantifiable guidance.
\newblock In \emph{Proceedings of the 2018 Conference on Empirical Methods in
  Natural Language Processing}, pages 3855--3864.

\bibitem[{Lin(2004)}]{lin2004rouge}
Chin-Yew Lin. 2004.
\newblock Rouge: A package for automatic evaluation of summaries.
\newblock \emph{Text Summarization Branches Out}.

\bibitem[{Ma et~al.(2018)Ma, Sun, Li, Li, Li, and Ren}]{ma2018query}
Shuming Ma, Xu~Sun, Wei Li, Sujian Li, Wenjie Li, and Xuancheng Ren. 2018.
\newblock Query and output: Generating words by querying distributed word
  representations for paraphrase generation.
\newblock In \emph{Proceedings of the 2018 Conference of the North American
  Chapter of the Association for Computational Linguistics: Human Language
  Technologies, Volume 1 (Long Papers)}, volume~1, pages 196--206.

\bibitem[{Och and Ney(2003)}]{och03:asc}
Franz~Josef Och and Hermann Ney. 2003.
\newblock A systematic comparison of various statistical alignment models.
\newblock \emph{Computational Linguistics}, 29(1):19--51.

\bibitem[{Papineni et~al.(2002)Papineni, Roukos, Ward, and
  Zhu}]{papineni2002bleu}
Kishore Papineni, Salim Roukos, Todd Ward, and Wei-Jing Zhu. 2002.
\newblock Bleu: a method for automatic evaluation of machine translation.
\newblock In \emph{Proceedings of the 40th annual meeting on association for
  computational linguistics}, pages 311--318. Association for Computational
  Linguistics.

\bibitem[{Paszke et~al.(2017)Paszke, Gross, Chintala, Chanan, Yang, DeVito,
  Lin, Desmaison, Antiga, and Lerer}]{paszke2017automatic}
Adam Paszke, Sam Gross, Soumith Chintala, Gregory Chanan, Edward Yang, Zachary
  DeVito, Zeming Lin, Alban Desmaison, Luca Antiga, and Adam Lerer. 2017.
\newblock Automatic differentiation in pytorch.
\newblock In \emph{International Conference on Learning Representations}.

\bibitem[{Prakash et~al.(2016)Prakash, Hasan, Lee, Datla, Qadir, Liu, and
  Farri}]{prakash2016neural}
Aaditya Prakash, Sadid~A Hasan, Kathy Lee, Vivek Datla, Ashequl Qadir, Joey
  Liu, and Oladimeji Farri. 2016.
\newblock Neural paraphrase generation with stacked residual lstm networks.
\newblock \emph{arXiv preprint arXiv:1610.03098}.

\bibitem[{See et~al.(2017)See, Liu, and Manning}]{see2017get}
Abigail See, Peter~J Liu, and Christopher~D Manning. 2017.
\newblock Get to the point: Summarization with pointer-generator networks.
\newblock In \emph{Proceedings of the 55th Annual Meeting of the Association
  for Computational Linguistics (Volume 1: Long Papers)}, pages 1073--1083.

\bibitem[{Su and Yan(2017)}]{su2017cross}
Yu~Su and Xifeng Yan. 2017.
\newblock Cross-domain semantic parsing via paraphrasing.
\newblock In \emph{Proceedings of the 2017 Conference on Empirical Methods in
  Natural Language Processing}, pages 1235--1246.

\bibitem[{Sun and Zhou(2012)}]{sun2012joint}
Hong Sun and Ming Zhou. 2012.
\newblock Joint learning of a dual smt system for paraphrase generation.
\newblock In \emph{Proceedings of the 50th Annual Meeting of the Association
  for Computational Linguistics: Short Papers-Volume 2}, pages 38--42.
  Association for Computational Linguistics.

\bibitem[{Sutton et~al.(2000)Sutton, McAllester, Singh, and
  Mansour}]{sutton2000policy}
Richard~S Sutton, David~A McAllester, Satinder~P Singh, and Yishay Mansour.
  2000.
\newblock Policy gradient methods for reinforcement learning with function
  approximation.
\newblock In \emph{Advances in neural information processing systems}, pages
  1057--1063.

\bibitem[{Vaswani et~al.(2017)Vaswani, Shazeer, Parmar, Uszkoreit, Jones,
  Gomez, Kaiser, and Polosukhin}]{vaswani2017attention}
Ashish Vaswani, Noam Shazeer, Niki Parmar, Jakob Uszkoreit, Llion Jones,
  Aidan~N Gomez, {\L}ukasz Kaiser, and Illia Polosukhin. 2017.
\newblock Attention is all you need.
\newblock In \emph{Advances in Neural Information Processing Systems}, pages
  5998--6008.

\bibitem[{Wang et~al.(2017{\natexlab{a}})Wang, Finch, Utiyama, and
  Sumita}]{wang2017sentence}
Rui Wang, Andrew Finch, Masao Utiyama, and Eiichiro Sumita. 2017{\natexlab{a}}.
\newblock Sentence embedding for neural machine translation domain adaptation.
\newblock In \emph{Proceedings of the 55th Annual Meeting of the Association
  for Computational Linguistics (Volume 2: Short Papers)}, pages 560--566.

\bibitem[{Wang et~al.(2017{\natexlab{b}})Wang, Utiyama, Liu, Chen, and
  Sumita}]{wang2017instance}
Rui Wang, Masao Utiyama, Lemao Liu, Kehai Chen, and Eiichiro Sumita.
  2017{\natexlab{b}}.
\newblock Instance weighting for neural machine translation domain adaptation.
\newblock In \emph{Proceedings of the 2017 Conference on Empirical Methods in
  Natural Language Processing}, pages 1482--1488.

\bibitem[{Wang et~al.(2018)Wang, Gupta, Chang, and Baldridge}]{wang2018task}
Su~Wang, Rahul Gupta, Nancy Chang, and Jason Baldridge. 2018.
\newblock A task in a suit and a tie: paraphrase generation with semantic
  augmentation.
\newblock \emph{arXiv preprint arXiv:1811.00119}.

\bibitem[{Williams(1992)}]{williams1992simple}
Ronald~J Williams. 1992.
\newblock Simple statistical gradient-following algorithms for connectionist
  reinforcement learning.
\newblock \emph{Machine learning}, 8(3-4):229--256.

\bibitem[{Wiseman et~al.(2018)Wiseman, Shieber, and Rush}]{wiseman2018learning}
Sam Wiseman, Stuart Shieber, and Alexander Rush. 2018.
\newblock Learning neural templates for text generation.
\newblock In \emph{Proceedings of the 2018 Conference on Empirical Methods in
  Natural Language Processing}, pages 3174--3187.

\end{thebibliography}
\bibliographystyle{acl_natbib}

\newpage

\newpage
\appendix

\section{Algorithm for extracting templates}
\begin{algorithm}
\caption{ExtractSentParaPattern}\label{euclid}
\begin{algorithmic}[1]
\INPUT $X$, $Y$, $Z^{\rmx}$, $Z^{\rmy}$, $\alpha'$, $V$
\OUTPUT $\bar{X}$, $\bar{Y}$
\Procedure{Extract $\bar{X}$}{}
\State $L \gets |X|$;
\State $\bar{X} \gets [\ ]$;
\State $c \gets 1$;
\State $p \gets  [\ ]$;
\For{$l:=1$ to $L$}
    \If {$z_{l}^{\rmx} = 0$}
        \If {$l=1$ \textbf{or} $\bar{X}_{l-1}\notin V$}
            \State $\bar{X}$.add($V_{c}$);
            \State $p$.add($[\ ]$);
            \State $c \gets c + 1$;
        \Else
            \State $p[c]$.add($l$);
        \EndIf
    \Else \State $\bar{X}$.add($X_{l}$);
    \EndIf
\EndFor
\EndProcedure
\Procedure{Extract $\bar{Y}$}{}
\State $T \gets |Y|$;
\State $\bar{Y} \gets [\ ]$;
\For{$t:=1$ to $T$}
    \If {$z_{t}^\rmy = 0$}
        \State $c \gets \underset{c}{\arg\max} \frac{1}{|p[c]|}\sum_{l=1}^{|p[c]|}\alpha_{p[c][l],t}'$;
        \If {$t=1$ \textbf{or} $\bar{Y}_{t-1} \ne V_{c}$}
            \State $\bar{Y}$.add($V_{c}$);
        \EndIf
    \Else
        \State $\bar{Y}$.add($Y_{t}$);
    \EndIf
\EndFor
\EndProcedure
\textbf{End}
\end{algorithmic}
\end{algorithm}

\begin{table*}[!htb]
    \centering
    \caption{Examples of the generated paraphrases and extracted patterns at each granularity level by DNPG.}\label{tab:gen-adap}
    \vspace{-5pt}
    \resizebox{1.0\linewidth}{!}{
    \begin{tabular}{p{5cm}p{5cm}p{5cm}p{5cm}p{5cm}p{5cm}}
        \toprule
        Input Sentence & Extracted Source Templates & Generated Sentential Paraphrase & Generated Paraphrase & Generated by MTL+Copy \\
        \midrule
        is there any verify angel investor on quora? & is there any \$x on \$y & how many \$x on \$y & how many verify angel investor on quora? & is there any verify on quora? \\

        how much salary do iit professor get? & how much salary do \$x get & how much money do \$x make & how much money do iit professor make? & how much do professor UNK?\\

        which is the best mobile below 15000? &  which is the \$x & the \$x & the best mobile below 15000 ? & what mobile 15000? \\

        how many time should i have bath? & how many \$x & number of \$x & number of time should i have bath? & how do you have bath?\\

        who is the best hollywood actor? & who is the \$x & what is \$x name & what is the best hollywood actor name? & who is the best actor? \\

        how do you change a key ignition in a 1988 chevy celebrity? & how do you \$x & what is the best way to \$x & what is the best way to change a key ignition in a 1988 chevy celebrity? & how do you change a 1988 in a 1988 chevy? \\

        why do company issue bonus share ? & why do \$x & what is the purpose of the \$x & what is the purpose of the company issue bonus share? & how do company issue bonus share? \\

        under which condition do the hiv virus survive? & under which condition do the \$x & which condition is best for the \$x & which condition is best for the hiv virus survive? & which do the hiv virus survive?\\

        use of monggo seed ? & \$x of \$y & what is the \$x of \$y & what is the use of monggo seed? & ? \\

        how do you eat potato salad? & how do you \$x & is there any way to \$x & is there any way to eat potato salad? & how do you eat potato salad? \\

        who is the most important person in yours life? & who is the \$x in \$y & who is \$y 's \$x & who is yours life 's most important person? & what is the most important person in yours life? \\

        what is the easiest business to start? & what is \$x & what is \$x in the world & what is the easiest business to start in the world? & what is business?\\

        \bottomrule
    \end{tabular}
    }
\end{table*}

\newpage
\section{Evaluation Guide}
Please evaluate the paraphrase with respect to three criterions: readability, accuracy, and diversity, which are arranged by importance. Specifically, the criterions of paraphrase quality from bad to good are listed in detailed as following:
\begin{enumerate}
  \item Non-readable. The generated paraphrase does not make sense and is not human-generated text. Please note that readable is not equivalent to grammatical correct. That is, considered there are non-English speaker, a readable paraphrase can have grammar mistakes.
  \item Readable but is not accurate. The answer to the paraphrased question is not helpful to the owner of the original question. For instance, \textit{how can i study c++} $\to$ \textit{what be c++}. Here are some examples of accurate paraphrase:
      \begin{enumerate}
        \item \textit{how can i learn c++} $\to$ \textit{what be the best way to learn c++}
        \item \textit{can i learn c++ in a easy way} $\to$ \textit{be learn c++ hard}
        \item \textit{do you have some suggestion of well design app} $\to$ \textit{what be some well design app name}
        \item \textit{be study hard} $\to$ \textit{how study hard}
      \end{enumerate}
  \item Accurate but with trivial paraphrasing. Just remove or add some stop words. For instance, \textit{why can trump win the president election} $\to$ \textit{why can trump win president election}
  \item Novel paraphrasing. More or loss, there is information loss of a non-trivial paraphrase. Thus, again, determine whether the paraphrase is equivalent to the original question from the perspective of question owner. Furthermore, it is not necessary for a non-trivial paraphrase contains rare paraphrasing pattern. For instance, maybe there is lot of paraphrase with the pattern \textit{what be \$name} $\to$ \textit{some interesting facts about \$name}. But it can still be considered as non-trivial paraphrase.
\end{enumerate}
There are some other things to be noted:
\begin{enumerate}
  \item There maybe special token, that is, [UNK] in the generated paraphrase. A generated paraphrase with [UNK] should generally have higher rank.
  \item The same paraphrase should have same ranking. Otherwise, please try your best to distinguish the quality of paraphrase.
  \item Please do Google search first when you see some strange word or phrase for better evaluation.
  \item Please note that all the words are stemmed and lower case. Just assume all the words are in their right form. For instance, \textit{what be you suggestion of some english movie} is equivalent to \textit{What are your suggestions of some English movies}.
\end{enumerate}
\end{document}